\documentclass[conference]{IEEEtran}
\IEEEoverridecommandlockouts
\usepackage{cite}
\usepackage{amsmath,amssymb,amsfonts}
\usepackage{algorithmic}
\usepackage{graphicx}
\usepackage{textcomp}
\usepackage{xcolor}
\usepackage{multirow}
\usepackage[figuresright]{rotating}
\usepackage{pifont}
\newcommand{\xmark}{\ding{55}}%

\def\BibTeX{{\rm B\kern-.05em{\sc i\kern-.025em b}\kern-.08em
    T\kern-.1667em\lower.7ex\hbox{E}\kern-.125emX}}
\begin{document}

\title{A Novel Audio-Visual Information Fusion System for Mental Disorders Detection}

\author{\IEEEauthorblockN{Yichun Li, Shuanglin Li, Syed Mohsen Naqvi}
\IEEEauthorblockA{Intelligent Sensing and Communications Research Group, Newcastle University, UK}
}

\maketitle

\begin{abstract}
Mental disorders are among the foremost contributors to the global healthcare challenge. 
Research indicates that timely diagnosis and intervention are vital in treating various mental disorders. However, the early somatization symptoms of certain mental disorders may not be immediately evident, often resulting in their oversight and misdiagnosis. Additionally, the traditional diagnosis methods incur high time and cost.
Deep learning methods based on fMRI and EEG have improved the efficiency of the mental disorder detection process. However, the cost of the equipment and trained staff are generally huge. Moreover, most systems are only trained for a specific mental disorder and are not general-purpose.
Recently, physiological studies have shown that there are some speech and facial-related symptoms in a few mental disorders (e.g., depression and ADHD). 
In this paper, we focus on the emotional expression features of mental disorders and introduce a multimodal mental disorder diagnosis system based on audio-visual information input. Our proposed system is based on spatial-temporal attention networks and innovative uses a less computationally intensive pre-train audio recognition network to fine-tune the video recognition module for better results. We also apply the unified system for multiple mental disorders (ADHD and depression) for the first time.
The proposed system achieves over 80\% accuracy on the real multimodal ADHD dataset and achieves state-of-the-art results on the depression dataset AVEC 2014. 

\end{abstract}


\section{Introduction}
Mental health encompasses an individual's psychological, emotional, and social well-being, which includes the ability to cope with stress, manage emotions, maintain relationships, and make decisions \cite{nash2022machine}. It plays a crucial role in overall health and functioning, influencing thoughts, feelings, and actions in daily life.
Fig. 1 shows the age (years) and gender distribution of patients with a diagnosis of severe mental illness (SMI) compared with all patients recorded by the United Kingdom National Health Service (NHS), UK. The results suggest that approximately 5-15 \% of recorded patient visits within the 20-60 years age group are impacted by severe mental illness, constituting a substantial overall figure. 
Various factors, including genetics, environment, life experiences, and biological factors, can influence mental health. As a result, mental disorders such as depression, Attention Deficit Hyperactivity Disorder (ADHD), and anxiety are common, particularly among children and adolescents \cite{chen2020adhd, sajjadian2021machine}. According to \cite{twostream}\cite{kelasi}, ADHD affects around 5-7 $\%$ of children and adolescents worldwide. Moreover, the global prevalence of depression was estimated at 28 $\%$ in 2021. These disorders could have serious consequences for individuals, including learning difficulties, impaired social interactions, and emotional issues \cite{ouyang2021evaluating,shuang2}.  

\begin{figure}[ht]
\centering
\includegraphics[scale=0.27]{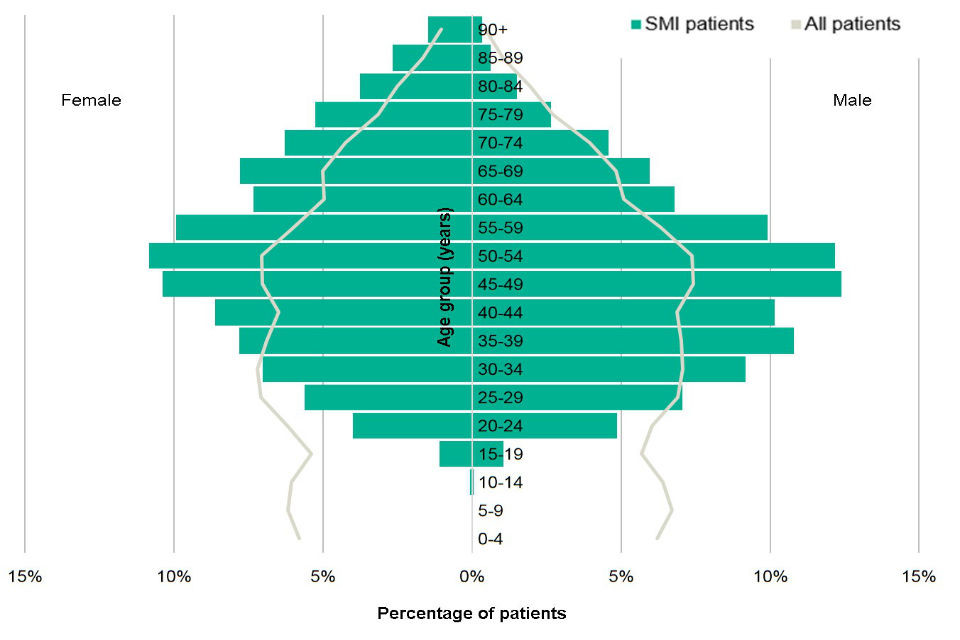}
\caption{\footnotesize Age (years) and gender distribution of patients with a diagnosis of severe mental illness (SMI) compared with all patients recorded by the United Kingdom National Health Service (NHS), UK \cite{UKNHS}. }
\end{figure}

The traditional diagnosis of mental disorders typically relies on clinicians' observation, questioning, and consultation, guided by the Diagnostic and Statistical Manual of Mental Disorders (DSM) \cite{edition2013diagnostic}. However, this diagnostic process is time-consuming and heavily dependent on the clinician's experience and judgment, including the long waiting time for the clinical appointment. The timely intervention significantly impacts the treatment of mental disorders and improves the quality of life for patients and their families \cite{yichun3,de2020encoding}. Due to the time-consuming process and the shortage of experienced clinical consultants, it has been reported that the waiting period for diagnoses and treatment of certain mental disorders such as ADHD, depression, and Alzheimer's can extend to several years \cite{nash2022machine}.
Recently, there has been a growing interest in machine learning methods for mental disorders detection and diagnosis. The majority of research in this area relies on Magnetic Resonance Imaging (MRI) and Electroencephalography (EEG) \cite{Lie2}. These methods efficiently detect and extract neurobiological symptoms and features, leveraging objective brain changes to diagnose subjects.
The conventional MRI and EEG-based methods have two limitations. Firstly, the expensive equipment and high operational costs limit the practical use of MRI and EEG in real-world diagnosis. MRI and EEG scanners cost £150,000-1,000,000 and £1,000-25,000, respectively, and the regular maintenance costs are also very expensive \cite{loh2022automated}. Secondly, recent detection and diagnosis techniques are specific to one mental disorder. However, many different mental disorders share the same or similar behavioral symptoms, such as dodge expression and uncontrollable body shaking, which are often overlooked by these neurobiological diagnostic techniques.

\begin{figure}[ht]
\centering
\includegraphics[scale=0.47]{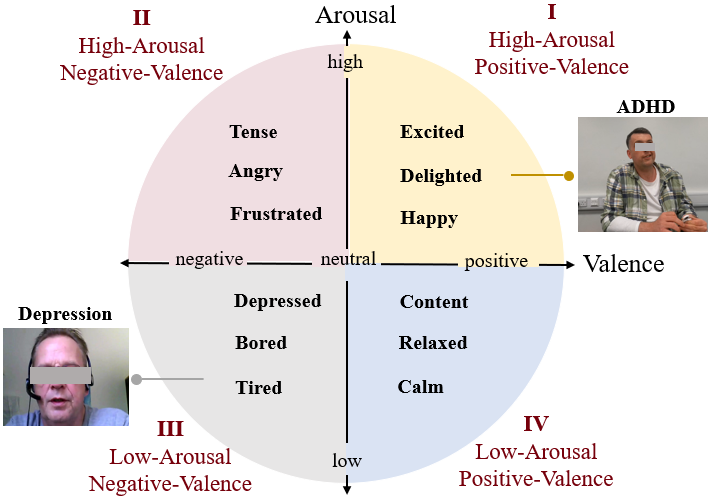}
\caption{\footnotesize The two-dimensional emotion space, which can be divided into four quadrants. Each quadrant is associated with various emotions. }
\end{figure}

Therefore, there is a growing demand for cost-effective and versatile psychiatric screening methods. In 1980, Russell introduced the concept of emotional states being represented as continuous numerical vectors in a two-dimensional space known as the Valence–Arousal (VA) space \cite{russell1989measures}. Valence denotes positive and negative emotional states, while arousal indicates the intensity of emotions ranging from sleepiness to high excitement. 
As shown in Fig. 2, depression typically occupies the third quadrant of the VA space, while ADHD is predominantly situated in the first and third quadrants. The distinct expression of various mental disorders within emotional space enables possible diagnosis and screening using a unified system. Many mental disorder symptoms manifest as observable emotional swings, which are reflected in both the patient's speech and facial expressions.


Therefore, we introduce a novel diagnostic system for mental disorders based on emotion recognition and the classification of audio and facial video input. 
The contributions of this paper are summarized as follows:

$\bullet $ A generalized diagnostic system for mental disorders is proposed, leveraging emotional recognition from raw RGB facial video and speech audio data. The performance of the system is also validated by the medical-approved NHS body.

$\bullet $ An efficient multimodal detection method is also proposed and applied to the mental disorder diagnosis for the first time. The innovative use of the simple pre-train audio model to fine-tune the video-based model to improve accuracy is provided.

$\bullet $ We demonstrate the effectiveness of our proposed method over both ADHD and depression datasets, compared to state-of-the-art benchmarks across the latest performance matrix.

The rest of the paper is organized as follows. The related work to mental disorders and diagnosis techniques is introduced in Section II. Then, the proposed method is described in Section III. The experimental settings and results are presented in Section IV. Finally, our work is concluded in Section V. It should be noted that this paper aims to explore the application of fusion systems based on audio-visual features in mental disorder assessment and detection. Further feature fusion experiments and more comprehensive detection results, will be addressed in the future journal paper of this work.

\section{Related Works}

Various physiological and psychological researchers have verified that lesions in certain brain areas can lead to behavior disorders. Therefore, for some psychological disorders, diagnosis and detection through emotional expression and behavioral characteristics have been proven feasible \cite{sharma2020impact,rissola2021survey,tang2022adhd2}.

Depression is a common psychiatric disorder. The DSM-V characterizes depression as enduring sadness and diminished interest in previously enjoyed activities. It also highlights that individuals might encounter supplementary physical symptoms, including chronic pain or digestive problems. In the research conducted by Li et al. \cite{li2023diagnosis}, they employed a methodology that calculates the two
attributes of brain regions based on the multi-layer network of dynamic functional connections and fuses morphological and
anatomical network features to diagnose depression, resulting in a reasonable classification accuracy of 93.6\%. In recent research, machine learning methods based on emotion recognition have also been used in depression assessment. In the research conducted by Niu et al. \cite{niu2023wavdepressionnet}, they proposed a representation block to find a set of basis
vectors to construct the optimal transformation space and generate the transformation result.

\begin{figure*}[h]
\centering
\includegraphics[scale=0.52]{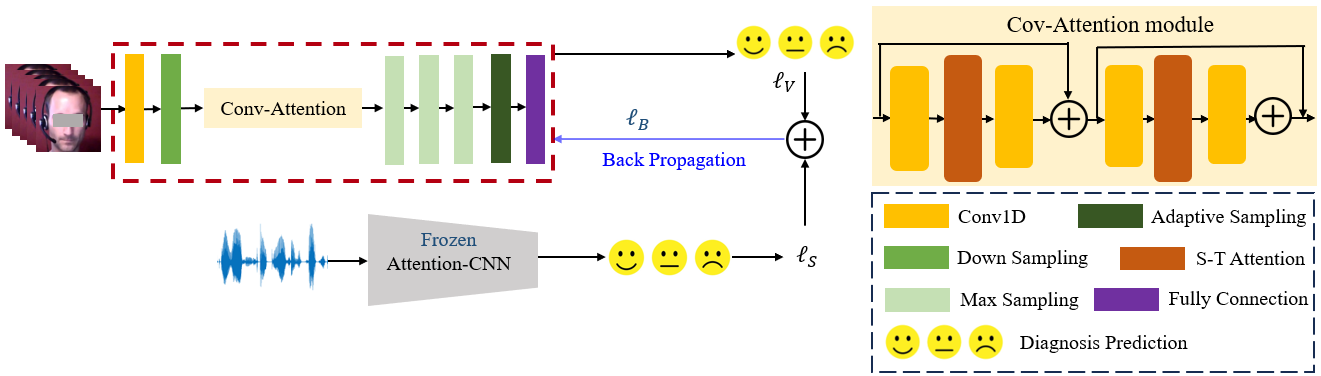}
\caption{\footnotesize Illustration of the proposed mental disorders assessment and detection system. The spectral-temporal feature from the video is extracted with the Cov-Attention module, 1D-convolutional layers, and spatial-temporal networks. We use a pre-train Attention-CNN module based on audio to fine-tune the video input. The $\oplus$ symbol denotes the concatenation operation for the fusion loss. The variables $\ell _{S}$, $\ell _{V}$, and $\ell _{B}$ denote the loss of the audio-based recognition model, the loss of the video-based recognition model and the fusion background loss, respectively. }
\end{figure*}

Brain MRI is the most widely used modality for ADHD diagnosis with machine learning. Most studies have sourced their MRI images from a single public database, namely, the Neuro Bureau ADHD-200 Reprocessed repository (ADHD-200) \cite{bellec2017neuro}. This dataset comprises structural and resting-state functional MRI images collected from 585 control individuals and 362 children and adolescents with ADHD. Numerous studies have identified structural differences between individuals with ADHD and controls. Based on this dataset, Tang et al. \cite{tang2022adhd3} achieve 99.6\% accuracy by employing a modified auto-encoder network to extract discriminative features and enlarge the variability scores for the binary comparison. 

Audio and video, as the most readily available multimodal signals, play a crucial role in various multimodal machine learning applications \cite{li2020single}. Their advantages include providing rich sensory information, enabling a better understanding of context, and facilitating natural interaction. Additionally, they greatly improve the performance and robustness of related systems by combining different features \cite{eason2}. 

Based on our investigation, the majority of mental disorders detection and diagnosis relies on fMRI and EEG tools, incurring high human and instrumental costs for application. Diagnostic and detection multimodal methods are highly limited, with most research on text and wearable sensors.
Therefore, developing a universal mental disorder detection system based on low-cost audio-visual signals has great potential for application.

\section{Proposed Methods}

This section briefly introduces the outlines of our systems and datasets utilized in this work. As our main emphasis lies on the multimodal fusion system based on audio and videos, details of networks utilized in the proposed system are also presented in this section.

\subsection{Proposed Multimodal Mental Disorder Detection systems}

The proposed system contains three main parts: video-based facial expression detection, audio-based pre-train model, and classification and regression performance measurements. 
Because of the particularity of medical and clinical-related information, open-source mental disorders datasets are relatively limited at this stage. We select multimodal datasets containing real facial video and audio encompassing a broad spectrum of mental disorders, including ADHD and depression.
Fig. 3 illustrates our proposed mental disorder detection system. Details will be introduced in the following subsections.

\subsection{Datasets}

This paper primarily focuses on attention deficit hyperactivity disorder (ADHD) and depression. These psychiatric conditions are selected due to their prevalence as the most common mental disorders, each characterized by specific emotional symptoms. Two challenging multimodal datasets, which serve as benchmarks for ADHD and depression detection and assessment, are utilized in the experiments. All datasets used to validate the detection and assessment performance of the proposed system are approved by certified medical authorities.

In our proposed system, we utilize the interview segments from our real multimodal ADHD dataset for the binary classification of ADHD \cite{yichun3}. Each subject and control undergoes a data recording process lasting 10-20 minutes, involving 21 questions selected from the Diagnostic Interview for ADHD in Adults (DIVA) administered in English. Notably, DIVA is a standard questionnaire used by doctors for ADHD diagnosis.
The recording setup involves three GoPro cameras: a front-facing Camera 1 captures facial information, while side Cameras 2 and 3 record the left and right torsos and limbs at a resolution of $3840\times2160$ pixels. For our proposed mental disorder assessment and detection system, only facial information captured by Camera 1 is utilized.
Videos are segmented into 60-second clips, and each is labeled as either 0 (non-ADHD controls) or 1 (ADHD subjects). The ADHD dataset used in our proposed system comprises 188 video clips, partitioned into training, validation, and testing at a ratio of 6:1:3, respectively. 

For depression detection, we employ the AVEC 2014 dataset to train and evaluate our proposed system. This dataset consists of videos accompanied by BDI-II score labels, self-evaluated by the participants in each video. The labels span from 0 to 63 and are divided into four depression levels: minimal (0–13), mild (14–19), moderate (20–28), and severe (29–63). The AVEC 2014 dataset consists of a total of 300 video recordings,
which are divided into three categories: training, development, and testing sets. Each set contains two different types of video recordings: Freeform and Northwind. The Northwind task involves participants reading aloud the fable `The North Wind and the Sun' in German. The Freeform task requires participants to answer a series of questions in German. The length of each video is approximately between 10 to 60 seconds.


\subsection{Networks}

As shown in Fig. 3, we design and present a two-stream multimodal system-based attention network with CNN and ResNet. The novelty is also in utilizing a pre-trained audio-based model to fine-tune the loss obtained from video input and get better results.

Based on the raw audio, we choose an attention-CNN structure as the main core network \cite{pan2024spatial}.
 A simple attention module is added to the CNN structure. Compared to traditional CNN, it focuses on leveraging local feature connections while utilizing parallel computing to decrease training time. The weights in the convolution kernels are shared, and multiple convolution kernels can be used to extract multi-dimensional information \cite{tran2015learning}.

There are 5 convolution layers that have 3$\times$3 kernels with 1 stride. 
Different from the original CNN audio classification network, we freeze the model after fully training the network. Then, we integrate this simple pre-train audio recognition model into the multimodal system to fine-tune the video recognition model.
 
The loss of the audio-based recognition model ($\ell _{S}$) is to minimize the Mean Absolute Error (MAE) of the outputs and true label results:
\begin{equation}
\ell_{S} = \frac{1}{n} \sum_{i = 1}^{n} \left | \hat{y}_{i} - y_{i} \right |
\end{equation}
where $n$ means the number of samples in the dataset, the $\hat{y}_{i}$ and $y_{i}$ are the predicted value and the true value of the $i$th sample, respectively.

For the video-based recognition and classification network, we proposed a Cov-Attention module based on the ResNet backbone. This module aims to capture both global and local spatial-temporal information from video frames.
The network architecture is illustrated in Fig. 3. 
In the first Cov1D layer, a convolution operation with a kernel size of 7$\times$7$\times$7 and a stride of 1x2x2 is applied to extract and downsample low-level features. Subsequently, a 3$\times$3$\times$3 pooling layer with a stride of 1$\times$2$\times$2 is employed to further process the features. The processed features are then passed through a residual module, which comprises two bottleneck structures.
Notably, the attention module replaces the middle layer of the bottleneck and generates a weighted feature incorporating attention features. Following a stack of residual modules, an adaptive pooling layer resamples the feature into a fixed shape. Finally, the last fully connected layer predicts a score, serving as the output of the proposed system.

The loss of the video-based recognition model ($\ell _{V}$) is to minimize the Mean Absolute Error (MAE) of the video outputs and true label results:
\begin{equation}
\ell_{V} = \frac{1}{n} \sum_{j = 1}^{n} \left| \hat{y}_{j} - y_{j} \right|
\end{equation}
where $n$ means the number of samples in the dataset. The $\hat{y}_{j}$ and $y_{j}$ are the predicted value and the true value of the $j$th sample, respectively.

We propose to perform loss fusion in this system, using the loss $\ell _{S}$ of an independent pre-train audio recognition model to fine-tune the loss $\ell _{V}$ on the video side to achieve more accurate recognition results. The fine-tune loss $\ell _{B}$ is defined as: 
\begin{equation}
\ell _{B}=\alpha \cdot  \ell _{S}+  \beta \cdot  \ell _{V}
\end{equation}
where the $\alpha$ and $\beta$ are the mixing coefficients to achieve the highest recognition effect through grid search and set empirically to 0.6 and 0.4, respectively.

Based on this basic system, we adapt two popular networks, i.e., LSTM and 3D-CNN \cite{tran2018closer}, to the video-based classification tasks. We also compare the proposed method with other commonly used audio-based recognition networks (LSTM) \cite{eason3}. To ensure the fairness of the evaluation, we use the same fusion method and parameter settings as in our proposed system.


\section{Experiments}

\subsection{Preprocessing and Experimental Settings}

To optimize the system's capacity to extract features from multimodal inputs, it is advantageous to perform preprocessing on the audio and video signals. By subjecting the preprocessing process, such as normalization, denoising, and feature extraction, the system can better discern relevant information from both modalities, thereby improving the performance of the system.

\begin{figure}[ht]
\centering
\includegraphics[scale=0.26]{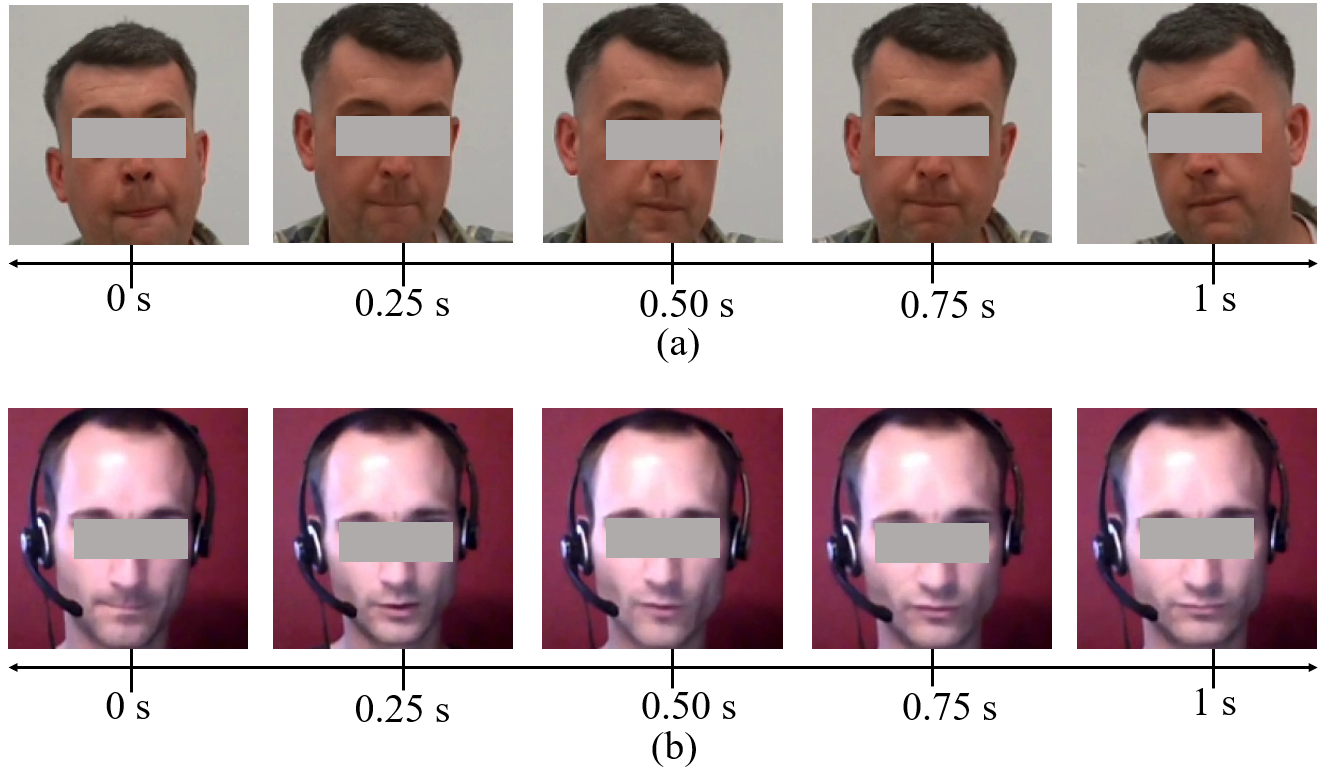}
\caption{\footnotesize Illustration of the facial reactions closely associated with mental disorders, i.e., ADHD (a) and depression (b), randomly selected subjects from multimodal ADHD data and AVEC 2014 data, respectively. }
\end{figure}

For optimal preservation of short-term facial expressions, we initially extract the raw video input as individual images at a 1-frame interval. Our preprocessing utilizes the Dlib toolkit to precisely extract facial landmarks from the sequence of frames, effectively eliminating background interference and aligning human faces to minimize environmental disruptions. As shown in Fig. 4,  during alignment, we center the facial features precisely between the eyes and adjust the vertical distance between the eyes and mouth to occupy one-third of the image's height. Subsequently, the aligned facial images are resized to the dimension of 224 × 224 pixels for further processing.


We extract audio data from the original recordings and proceed to eliminate background noise from the extracted audio samples. This process is achieved using the noisereduce() function in Python.
Fig. 5 shows the audio spectrograms separated by denoising from the original recordings. The noise threshold is determined based on statistical analysis performed across the audio clip. 
Following the denoising process, the raw audios are segmented into clips of a 2-second duration and organized within the same structure as the corresponding video dataset.

\begin{figure}[ht]
\centering
\includegraphics[scale=0.6]{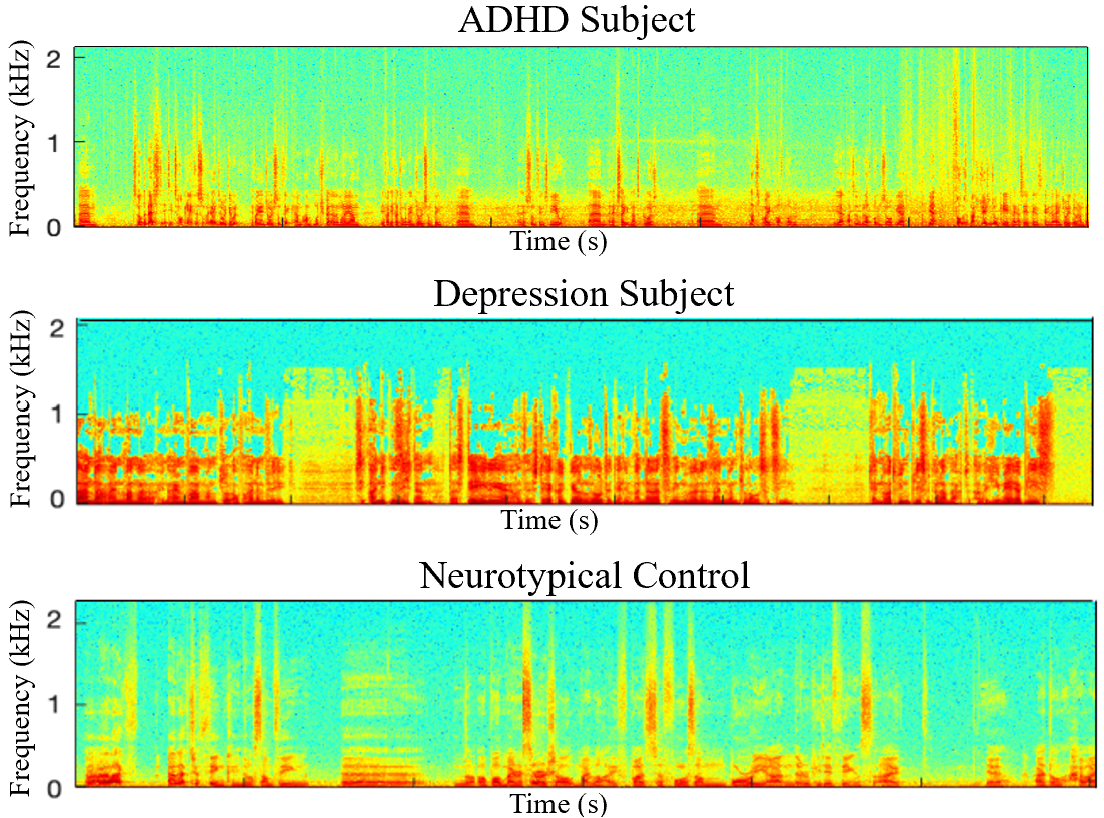}
\caption{\footnotesize Illustration of the audio spectrogram from a randomly selected ADHD subject, depression subject, and neurotypical control. }
\end{figure}

Our proposed system is trained on the AVEC 2014 and the real multimodal ADHD dataset. We subsequently validate and test it on the development/validation and testing sets of both datasets, respectively.
For the training period for the multimodal system, we randomly select a sequence of 64 frames from a given audio-visual recording at a stochastic position to form a training clip. To augment the input data during training, we apply random horizontal flips and adjust brightness, contrast, saturation, and hue within the range of 0 to 0.1 for all frames in one video clip.
During the testing phase, for a given testing audio-visual recording, we crop it into a sub-video with 64 frames and compare it with the indicated audio group, predicting each group to calculate the mean value of depression scores and the ADHD classification probability score. The training epochs for the audio model is 100, with the learning rate empirically set to $1\times 10^{-4}$. The training epochs for the multimodal network are 150, and the learning rate is empirically set to $1\times 10^{-3}$. All experiments are run on a workstation with four Nvidia GTX 1080Ti GPUs.

 \subsection{Results and Discussions}

As mentioned in Section III, our proposed system differs from traditional machine learning approaches based on fMRI and EEG. Instead, it concentrates on creating multimodal systems for detecting and assessing mental disorders using video and audio data collected by common sensors. These multimodal systems have the advantage of correcting or compensating for errors in detection that can arise from single-modality inputs, making them especially valuable for such related medical research.

Our proposed system's performance is evaluated by comparing it with LSTM and 3D-CNN networks on this real multimodal ADHD dataset with precision, accuracy, and Area Under the Curve (AUC) to evaluate the classification performance of the system.
For the results of the AVEC 2014 dataset for depression, we utilize Mean Absolute Error (MAE) and Root Mean Squared Error (RMSE). 
The experimental results are summarized in Table I and Table II. 

\begin{table}[ht]
\caption{\footnotesize ADHD detection performance with different neural networks. \textbf{Bold} indicates the best results. \\ }
\centering
\vspace{-1em}
\begin{tabular}{c|c| c c c c }
\hline
\textbf{Audio}&\textbf{Video}&\textbf{Precision}&\textbf{Accuracy}&\textbf{AUC}\\ \hline
LSTM& 3D CNN &75.58&59.60&64.84\\ \hline
LSTM& LSTM &65.67&47.98&53.13\\ \hline
Attention CNN& 3D CNN &76.19&65.16&68.38\\ \hline
Attention CNN& LSTM &71.71&70.57&67.20\\ \hline
\textbf{Attention CNN}&\textbf{Cov-Attention} &\textbf{81.08}&\textbf{82.22}&\textbf{77.35}\\ \hline
\end{tabular}
\end{table}

\begin{table}[ht]
\caption{\footnotesize Depression detection performance with different neural networks. \textbf{Bold} indicates the best results. \\ }
\centering
\vspace{-1em}
\begin{tabular}{c|c| c c }
\hline
\textbf{Audio}&\textbf{Video}&\textbf{MAE}&\textbf{RMSE} \\ \hline
LSTM& 3D-CNN &8.81&10.04\\ \hline
LSTM& LSTM &10.43&12.11\\ \hline
Attention CNN& 3D CNN &7.67&9.24\\ \hline
Attention CNN& LSTM &8.52&10.20\\ \hline
\textbf{Attention CNN}& \textbf{Cov-Attention} &\textbf{7.23}&\textbf{9.36}\\ \hline
\end{tabular}
\end{table}

From Table I, the proposed system shows good classification ability on the real multimodal ADHD  dataset. At the same time, the results from Table II show that the proposed audio-visual attention network is significantly higher than the other fusion networks in MAE and RMSE. 

We also provided part of state-of-the-art results compared to our proposed method in Table III and Table IV. It should be emphasized that, due to medical confidentiality requirements, there is no publicly available ADHD multimodal dataset. Therefore, we evaluated the performance of state-of-the-art ADHD detection systems on various datasets containing EEG and daily activities videos. 

\begin{table}[h]
\caption{\footnotesize Comparison with STATE-OF-THE-ART METHODS for ADHD Detection.\\ }
\centering
\scriptsize\addtolength{\tabcolsep}{-4pt}
\begin{tabular}{c|c c c c }
\hline
\textbf{Author} & \textbf{Data Input} & \textbf{Classifier} & \textbf{Accuracy} & \textbf{Equipment Price (\$)} \\ \hline
\emph{Luo et al.\cite{luo2020multimodal}} & MRI \& DTI & CNN & 76.6\% & 150,000-1,000,000 \\ \hline
\emph{Peng et al.\cite{peng2021efficacy}} & fMRI & CNN & 72.9\% & 150,000-1,000,000 \\ \hline
\emph{Vahid et al.\cite{vahid2019deep}} & EEG & CNN & 80.3\% & 1,000-25,000 \\ \hline
\textbf{\emph{Proposed method}} & Audio-Visual & Cov-attention & \textbf{82.22\%} & \textbf{450} \\ \hline
\end{tabular}
\end{table}

\begin{table}[ht]
\caption{\footnotesize Comparison with state-of-the-art methods for Depression Assessment. \\ }
\centering
\vspace{-1em}
\begin{tabular}{c|c| c  }
\hline
\textbf{Author}&\textbf{Features}&\textbf{MAE}\\ \hline
\emph{Valstar et al.\cite{valstar2014avec}}& Hand-craft features &10.03\\ \hline
\emph{Niu et al.\cite{niu2020multimodal}}& Fourier
spectrogram &7.65\\ \hline
\emph{Niu et al.\cite{niu2019automatic}}& Mel spectrogram &7.67\\ \hline
\emph{Du et al.\cite{du2019encoding}}& Video &7.28\\ \hline
\textbf{Proposed method}& \textbf{Audio-Visual Information} &\textbf{7.23}\\ \hline
\end{tabular}
\end{table}

The proposed audio-visual fusion emphasizes extracting emotional information, specifically relevant symptom features of mental disorders, from both audio and video inputs. In this system, the Cov-Attention model captures expression and emotional cues across multiple contiguous frames, utilizing both spatial and temporal dimensions of video data. This model is crucial for accurate diagnosis and classification in related tasks. Additionally, the attention-CNN model in the audio recognition module effectively captures frequency and dimensional features present in speech signals,i.e., enhancing performance in speech-based detection and diagnosis while introducing the pre-train model. Overall, the proposed system exhibits strong performance in the detection tasks of the ADHD multimodal dataset, achieving high accuracy using only cost-effective audio-visual information data. Additionally, it demonstrates robust performance in the assessment of depression using the AVEC 2014 dataset.

We conduct ablation studies to assess the diagnostic performance of each module within our proposed system on the same AVEC 2014 dataset and multimodal ADHD dataset. The corresponding results are presented in Table V and Fig. 6, respectively.

\begin{table}[htbp!]
\caption{Ablation study of four contributions in the proposed method. \textbf{Bold} indicates the best results.}
\vspace{-1em}
\centering
\begin{tabular}{cccc}
\hline
\multicolumn{2}{c}{Ablation Settings} & %
    \multirow{2}{*}{Backbone}&\multirow{2}{*}{MAE}   \\
\cline{1-2}
   Audio & Video & \\
\hline
\checkmark &\xmark & Attention-CNN & 7.64\\
\xmark &\checkmark & Cov-Attention & 7.58\\
\hline 
\checkmark &\checkmark & \textbf{Attention+Cov-Attention} & 7.23\\
\hline 
\end{tabular}
\end{table}

\begin{figure}[ht]
\centering
\includegraphics[scale=0.475]{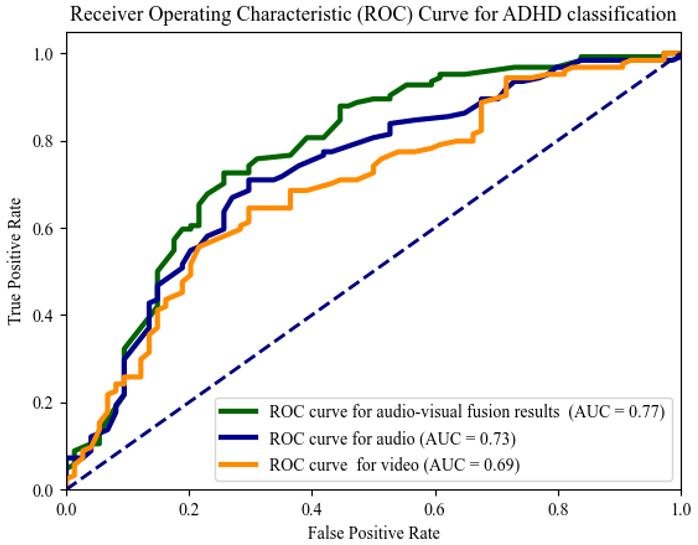}
\caption{\footnotesize ROC curve for ADHD classification ablation study result. }
\end{figure}

Based on the results of the ablation studies,  firstly, both the audio and video-based classification networks exhibit a notable level of robustness in assessing and detecting various psychological and emotional features, underscoring their efficacy in cross-mental disorders.
Secondly, fine-tuning the video model using simple pre-trained audio models leads to significant improvements in classification accuracy and performance across different experiments.
Thirdly, by leveraging the strengths of both network models and exploiting the features from different modalities, we have developed a comprehensive system for assessing and detecting various mental disorders. This integrated approach yields superior performance, particularly evident in depression assessment, where the MAE on the AVEC 2014 dataset is 7.23, and the AUC on the ADHD dataset is 0.77.

The aforementioned results highlight the similar related symptoms in emotional expression, including facial expressions and speech, across different mental disorders. They also indicate the feasibility of evaluating and screening multiple mental disorders through a unified multimodal system.
Moreover, while video-based depression assessment performs slightly better than only audio-based depression assessment, the opposite is observed for ADHD diagnosis. This discrepancy may be attributed to differences in symptom manifestation among various mental disorders. We note that the ADHD data are collected from interview videos, potentially amplifying the prominence of speech characteristics over facial expressions. 
In our future work, we intend to delve deeper into these findings through more comprehensive experiments and introduce more multimodal data, such as EEG and fMRI, for fusion and evaluation of the related disorders.

\section{Conclusion}
This paper presented an innovative multimodal detection system for identifying and detecting various mental disorders. The proposed system demonstrated state-of-the-art assessment and classification capabilities on depression and ADHD datasets, respectively. By leveraging a simple pre-train audio model to fine-tune video data, our system achieved promising results, as evidenced by comparative and ablation study experiments. Compared to conventional machine learning methods based on EEG and fMRI, our system offered cost-effectiveness and broader applicability, pointing to a promising direction in clinical practice.
For future research, we aim to broaden the scope of our proposed system to encompass a wider array of mental disorders with a larger sample size. It should be noted that this paper aims to explore the application of fusion systems based on audio-visual features in mental disorder assessment and detection. Further feature fusion experiments and more comprehensive detection results will be addressed in this work's future journal paper.

\section*{Acknowledgments}
We would like to express our gratitude to Dr. Rejesh Nair from the United Kingdom National Health Service (NHS), UK, for his professional medical advice and help and all participants and volunteers for the multimodal ADHD data recording. Especially the Cumbria, Northumberland, Tyne, and Wear (CNTW) NHS Foundation Trust, one of the largest mental health and disability Trusts in England.

\bibliographystyle{IEEEtran}
\bibliography{output}

\end{document}